# Comparing Performance of Preprocessing Techniques for Traffic Sign Recognition Using a HOG-SVM

Luis Vieira

*School of Mathematical and Computational Sciences*
*Massey University – Albany Campus,*
*Auckland, New Zealand*
23012096@massey.ac.nz

*June, 2024*

**Abstract**. This study compares the performance of various preprocessing techniques for Traffic Sign Recognition (TSR) using Histogram of Oriented Gradients (HOG) and Support Vector Machine (SVM) on the German Traffic Sign Recognition Benchmark (GTSRB) dataset. Techniques such as CLAHE, HUE, and YUV were evaluated for their impact on classification accuracy. Results indicate that YUV in particular significantly enhance the performance of the HOG-SVM classifier (improving accuracy from 89.65% to 91.25%), providing insights into improvements for preprocessing pipeline of TSR applications.

## 1 Introduction

Traffic Sign Recognition (TSR) is fundamental for Advanced Driver Assistance Systems (ADAS) and autonomous vehicles, ensuring road safety and efficient navigation. Accurate detection and classification of traffic signs are essential for these systems, requiring robust algorithms and effective preprocessing techniques.

Traditional machine learning methods, like Histogram of Oriented Gradients (HOG) combined with Support Vector Machine (SVM), are widely used in TSR for their robustness and computational efficiency. HOG captures essential shape and edge information, while SVM provides reliable classification performance in high-dimensional spaces.

The German Traffic Sign Recognition Benchmark (GTSRB) dataset, with over 50,000 images across 43 classes, is a key resource for evaluating TSR techniques. While deep learning methods show high accuracy, their computational demands make traditional approaches like HOG-SVM preferable for real time applications.

This study aims to evaluate the preprocessing pipeline for a HOG-SVM by comparing the performance of techniques such as CLAHE (Contrast Limited Adaptive Histogram Equalization), HUE, and YUV. By assessing their impact on classification accuracy, the research seeks to enhance TSR performance and contribute to more efficient and accurate systems.

# 2 Literature Review

With the increasing demand for ADAS and autonomous driving, TSR has become a focus area of research. In recent years there is an increased need for the ability to accurately detect and classify traffic signs ensuring road safety and efficient navigation. This literature review provides an overview over a wide range of methodologies and datasets in TSR field, from traditional machine learning techniques, deep learning advancements, and comprehensive benchmark datasets.

## 2.1 Traditional Machine Learning Approaches

Traditional machine learning methods have for long been the foundation for TSR systems, and even with the continuous advances in the field, they still provide quick reliable results. These methods include various feature extraction techniques and classifiers. Fleyeh and Roch (2013) evaluated HOG descriptors for classifying the Swedish speed limit signs using a Gentle AdaBoost classifier, achieving a high classification rate of 99.42%. The study confirms the robustness of SVMs and HOG in TSR applications.

Soni et al. (2019) combined Principal Component Analysis (PCA) features with SVM classifiers in order to improve recognition of signs on the Chinese Traffic Sign Database (TSRD), demonstrating the effectiveness of dimensionality reduction in enhancing the performance.

HAAR features by Viola and Jones (2001) have been broadly used in object detection tasks. They use simple rectangular features and a cascade classifier for quick detection, which makes them suitable for real time applications.

## 2.2 Deep Learning Approaches

Deep learning, especially Convolutional Neural Networks (CNNs) have been advancing in recent years and allowing automatic feature extraction and improving TSR accuracy.

Ciresan et al. (2012) developed a multi-column deep neural network (MCDNN) that achieved a recognition accuracy of 99.46% on the GTSRB dataset, outperforming traditional methods. Arcos-García et al. (2018) enhanced CNNs with spatial transformer networks (STNs) and stochastic gradient descent optimisation, achieving a recognition accuracy of 99.71% on the GTSRB, highlighting the benefits of incorporating spatial transformations and advanced optimisation techniques.

A real time TSR method using an attention-mechanisms based deep CNN was proposed by Triki et al. (2023), achieving a testing accuracy of 99.91% on the GTSRB dataset. This mechanism allowed the model to focus on significant image parts improving recognition performance.

## 2.3 Advanced Methods

Recent studies have explored new ways to enhance TSR performance that focus on computational efficiency and real time applicability.

A lightweight CNN architecture optimised for traffic sign recognition in urban environments was developed by Khan, Park, and Chae (2023). Their model achieved high accuracy rates on the GTSRB and Belgium Traffic Sign (BelgiumTS) datasets, demonstrating suitability for resource-constrained settings.

Khan and Park (2024) integrated Gradient-weighted Class Activation Mapping (Grad-CAM) and Local Interpretable Model-Agnostic Explanations (LIME) into their CNN-based TSR model, providing transparency and interpretability to address the black-box nature of deep learning models.

Jiawei Xing (2021) compared Faster R-CNN and YOLOv5 for TSR, concluding that YOLOv5 is more suitable for real time applications due to its higher accuracy and speed. The study also introduced image defogging methods to enhance TSR. Huang et al. (2014) proposed an ELM-based TSR method using HOG descriptors, demonstrating high recognition accuracy and computational efficiency on the GTSRB dataset. While Vega et al. (2012) introduced a method for detecting and reconstructing road signs using colour segmentation and Gielis curves, achieving an accurate detection rate of 81.01% on a dataset of 130 images.

## 2.4  Datasets

The quality and diversity of benchmark datasets play an important part for training and evaluating TSR models. The GTSRB dataset, organised by Stallkamp et al. (2012), includes over 50,000 images across 43 classes. It has become a standard benchmark for TSR research due to its comprehensive annotations and varied conditions.
Another popular dataset include the large Mapillary Traffic Sign Dataset (MTSD) introduced by Ertler et al. (2020), a diverse dataset with over 100,000 images from various global locations at street level. Other datasets include the Belgium, Swedish, Chinese and Austrian. The latter (Austrian Highway Traffic Sign Data Set (ATSD)) was presented by Maletzky et al. (2023) which includes around 7,500 scene images and over 28,000 annotated traffic signs, providing a rich dataset for training deep learning models in TSR.

The evolution of TSR systems from traditional machine learning to advanced deep learning techniques has significantly enhanced the accuracy and robustness of traffic sign recognition. Benchmark datasets like GTSRB and MTSD have played a pivotal role in driving these advancements. Future research should focus on further improving real time processing capabilities.

# 3  Methodology

## 3.1  SVM vs CNN

In TSR, SVMs and CNN are the two most commonly used methods. SVMs, used with feature extractors like HOG are popular for being robust dealing with high dimensional spaces and effective in classification tasks, while providing reliable performance with lower computational demands. Conversely, CNNs, which automatically learn feature hierarchies from raw images, have shown superior accuracy due to their deep learning capacity in capturing complex patterns. However, CNNs demand substantial computational power and large datasets for training, which may not be feasible in resource constrained setups.
This study opted for SVM due to its robust performance and efficiency, given the limited computational resources and the focus on evaluating preprocessing techniques rather than optimising deep learning models.

## 3.2  Use of HOG in TSR

In the field of TSR, selecting an appropriate feature extraction method is crucial for balancing accuracy with computational efficiency. HOG combined with a SVM, provides high performance in object detection tasks, is a standout feature descriptor due to its robustness, simplicity, and computational efficiency, making it an optimal choice for TSR under constrained resources.
HOG focuses on gradient structures rather than colour or intensity, making it robust to diverse lighting and weather conditions. It captures essential shape and edge information consistently, ensuring high accuracy in varied environments (Dalal & Triggs, 2005).
HOG's simplicity and efficiency are significant for TSR applications, in particular for real time. Basic operations like gradient computation and histograms are less demanding, making HOG suitable for embedded systems with limited processing power, common in real time automotive applications (Dalal & Triggs, 2005).

In contrast, alternative methods like SIFT (Scale-Invariant Feature Transform) and SURF (Speeded-Up Robust Features) whilst robust to scale and rotation, are computationally intensive, require more resources and longer processing times (Lowe, 2004; Bay et al., 2006). LBP (Local Binary Patterns) is efficient but less robust to lighting changes, often better for texture classification than object recognition (Ojala et al., 1996). HAAR features, on the other hand, although used widely for real time object detection due to its speed and effectiveness (Viola and Jones, 2001), is less robust to variations in lighting and pose, which can impact their performance in TSR.

In short, HOG is the most practical choice for TSR when considering computational constraints without the need for maximum performance.

## 3.3  Preprocessing Techniques

In this project, several preprocessing techniques were employed to enhance the performance of the HOG-SVM classifier for Traffic Sign Recognition (TSR). The primary techniques used include CLAHE, HUE, and YUV. Additionally, the discussion will contrast these techniques with other commonly used methods in TSR such as PCA and LDA, which were not employed in this project.

Gaussian Blur is used to reduce image noise and its detail by smoothening the image averaging the pixel values, which helps focusing on essential features of the traffic signs while reducing the effect of noise (Gonzalez & Woods, 2002). This preprocessing step enhances the ability of HOG to capture the edges and gradients effectively, contributing to better feature extraction and classification accuracy.

CLAHE (Contrast Limited Adaptive Histogram Equalization) is a technique used to improve the local contrast of an image. Operates on small regions (tiles) and applies histogram equalisation. It limits contrast enhancement in homogeneous areas to prevent noise amplification and combines results using bilinear interpolation to avoid visible boundaries (Zuiderveld, 1994). This method is useful in TSR for identifying features in images with poor lighting or low contrast.

HUE describes pure colours in the HSV and HSL colour spaces, measured as an angle on a colour wheel from 0 to 360 degrees. By focusing on hue, colour information is separated from intensity, which aids in object recognition, image segmentation, and colour based feature extraction (Smith, 1978). Hue adjustments enhance image features without affecting brightness, making it valuable in TSR.

YUV separates an image into luminance (Y) and chrominance (U and V) components, allowing for better compression efficiency and can handle colour in schemes like JPEG or MPEG (Poynton, 1998). In TSR, YUV is can be advantageous for separating colour information from intensity, enhancing feature extraction for edges and textures without affecting overall colour balance.

### 3.3.1 Other Commonly Used Techniques in TSR

The use of dimensionality reduction techniques like Linear Discriminant Analysis (LDA) and PCA Principal Component Analysis (PCA) have been shown to improve the TSR classification. Dimensionality reduction techniques like PCA and LDA improve TSR classification by reducing feature complexity, improving accuracy and computational efficiency. PCA maximizes variance, thus can aid in managing high-dimensional HOG features and enhancing SVMs performance. LDA maximizes class separability, optimising feature space for better classification accuracy. Integrating these techniques addresses high-dimensional data issues, improving real time performance (Jolliffe, 2002; Fisher, 1936).

Compared to other techniques like histogram equalisation, edge detection (e.g.: Canny), and normalisation, the methods used offer higher advantages. CLAHE is more advanced than basic histogram equalisation, and the combination of HUE and YUV provides a robust approach to colour handling. Techniques like PCA and LDA, while not used in this project, offer dimensionality reduction benefits that could be explored in future work to further improve performance.

### 3.4 Dataset

Figure 1: German road signs present in the GTSRB dataset (43 classes).

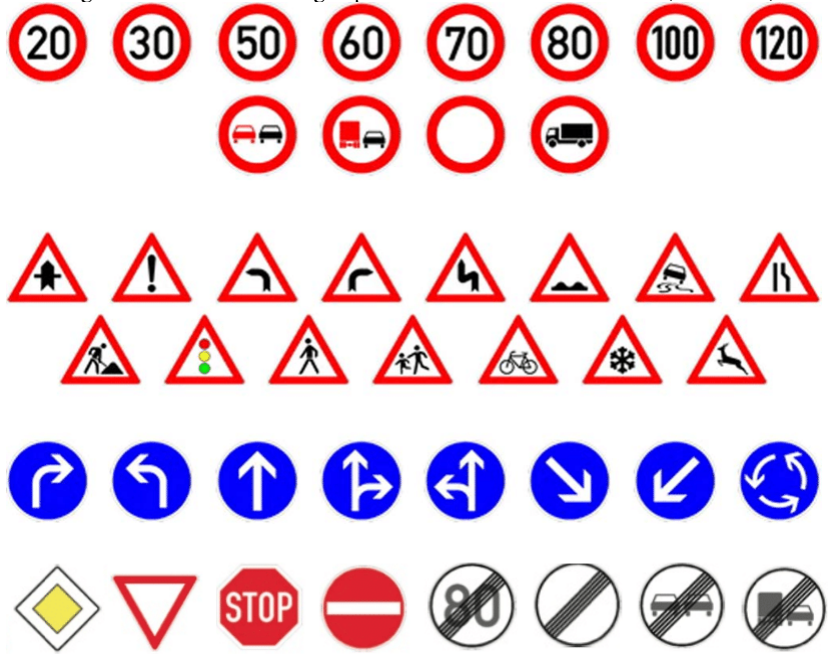

Source: http://dx.doi.org/10.1109/SII.2017.8279326

The GTSRB dataset used in this study contains over 50,000 images amongst 43 different classes (Figure 1), making it one of the most comprehensive resources for evaluating TSR techniques. Each image include one traffic sign each and include around the actual traffic sign a border of 10% that allows for edge based approaches. The image sizes found are between 15x15 and 250x250 pixels, while the actual traffic sign is often not centred in the image. The bounding box (ROI metadata – Figure 2) of the traffic sign is part of the annotations.

Figure 2: Sample ROI

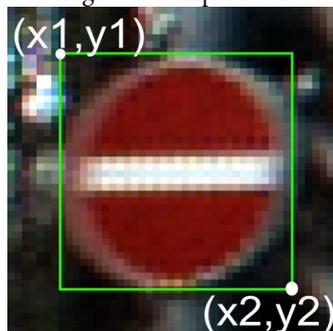

Source: https://benchmark.ini.rub.de/

## 3.5 Preprocessing Technical Procedure

HOG: Gaussian Blur was used to reduce image noise and detail, smoothing the image by averaging pixel values (kernel size 3x3, sigma = 0), before extracting features using HOG (window size: 32, block size: 16, stride: 8, cell size: 8, bins: 9). This was considered as the benchmark baseline for comparison with the other approaches.

For the CLAHE preprocessing, a dynamic adjustment was applied based on the standard deviation of the image's luminance channel, as follows:

. Low contrast (std dev < 50): clip limit = 4.0
. Medium contrast (std dev < 100): clip limit = 2.0
. High contrast (std dev >= 100): clip limit = 1.0
. Tile Grid Size: 8x8

This dynamic adjustment ensures that contrast improvement is correctly applied based on the initial image contrast, minimising noise amplification in low-contrast regions while enhancing for high-contrast areas.

CLAHE-HOG: Enhances image contrast using a dynamic CLAHE before applying HOG to improve feature quality.

YUV-HOG: Converts images to YUV colour space to separate luminance (Y) from chrominance (U and V), enhancing HOG's ability to capture essential features.

HUE-HOG: Focuses on the hue component, equalising it to improve feature extraction before applying HOG.

CLAHE-YUV-HOG: Sequentially applies CLAHE and YUV transformation before HOG, combining the benefits of contrast enhancement and colour space separation.

HUE-YUV-HOG: Extracts and equalises the hue, then converts to YUV space before applying HOG, aiming for robust feature extraction.

CLAHE-HUE-YUV-HOG: Combines all preprocessing techniques in sequence before applying HOG, hoping to maximise feature quality.

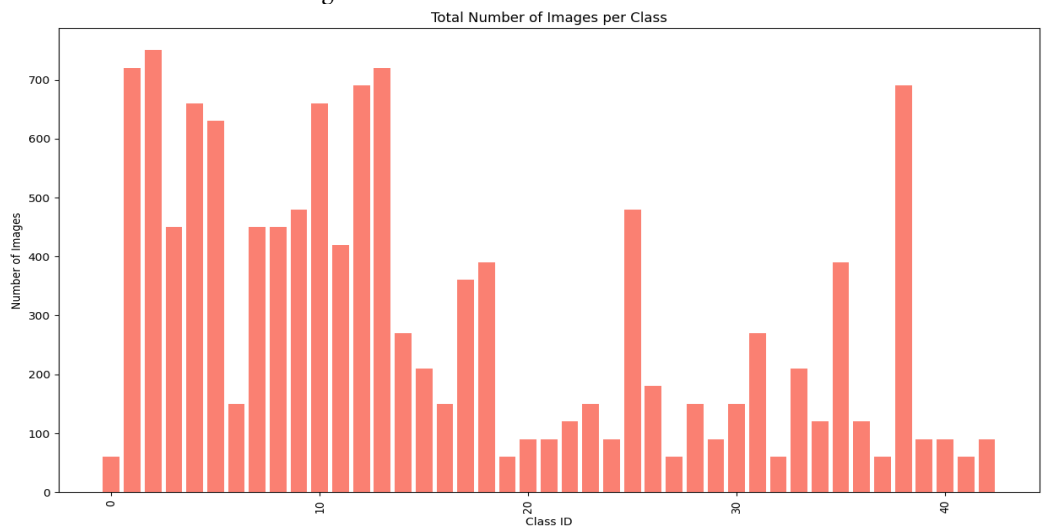

Figure 3: GTSRB Test Dataset Class Imbalance

The dataset was initially shuffled to reduce class imbalance (Figure 3) and minimise bias and overfitting, followed by a split into a training set and a validation set in an 80:20 ratio. Images were resized to a uniform dimension of 32x32 pixels to standardise the input for feature extraction.

Luis Vieira

Various preprocessing techniques (HOG with Gaussian Blur, CLAHE-HOG, YUV-HOG, HUE-HOG, CLAHE-YUV-HOG, HUE-YUV-HOG, and CLAHE-HUE-YUV-HOG) were applied to enhance and extract relevant features from the images. The features' extraction from all preprocessed images was done using HOG with Gaussian Blur. Then, the SVM was trained on the extracted features using the training set. Finally, the trained models were evaluated on validation and test sets to measure their performance.

## 3.6 Classifier

A Support Vector Machine (SVM) with an RBF kernel was used due to its effectiveness in classification tasks. The hyperparameters were optimised using C=20.5557 and γ=0.2167. The procedure to obtain these values involved a RandomizedSearchCV with an initial 5-fold cross-validation and 10 iterations over a wider range (0.5 < C < 50 and 0.01 < γ < 1, 10 samples in both cases), followed by a narrower range using 3-fold cross-validation and 10 iterations (5 < C < 25 and 0.05 < γ < 0.35, and 10 samples).

# 4 Results

In the context of our TSR using the GTSRB dataset, the chosen techniques (CLAHE, HUE, YUV, and Gaussian Blur) are clearly effective. CLAHE enhances local contrast, making traffic signs more distinguishable. HUE separates colour information from intensity, improving colour-based feature extraction. YUV allows for better handling of varying lighting conditions. Gaussian Blur reduces noise allowing for improved edge detection for HOG features.
It is a very challenging dataset and becomes clear that even humans would likely not be able to recognise every road sign accurately given the various distortions the images suffer. Figure 2 shows a random sample image from the first 5 classes visually comparing the original resized images with the various preprocessing techniques employed.

Figure 4: Output comparison of sample original images from 5 classes and the preprocessing techniques.

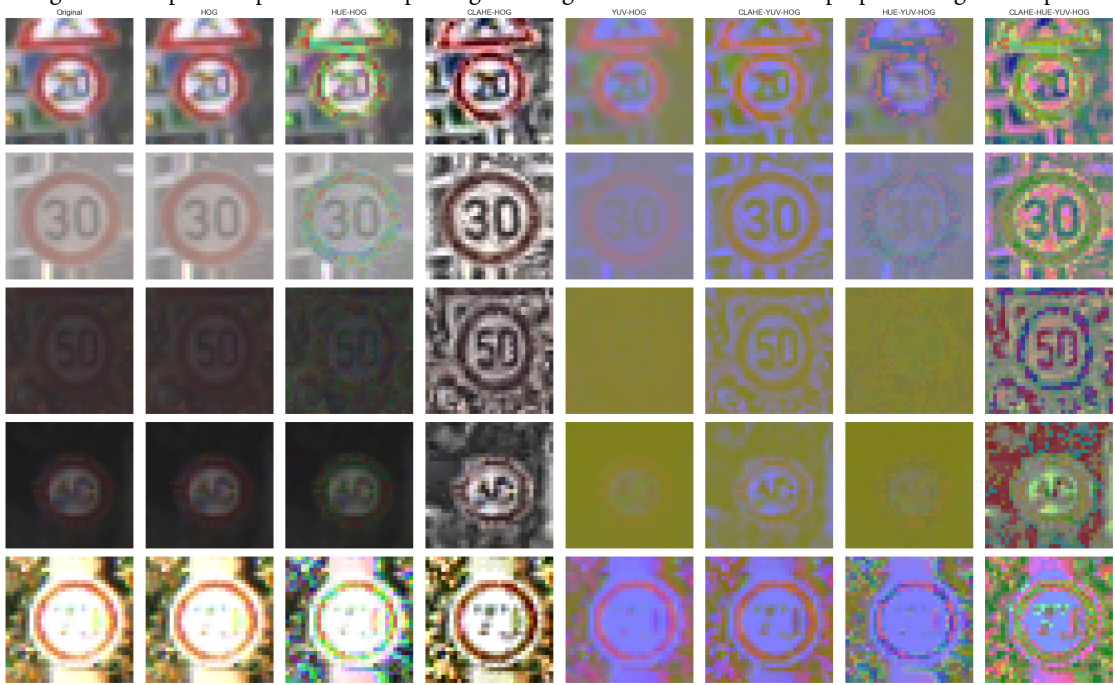

Nonetheless of the preprocessing techniques all show good classification performance, as found on tables 1 and 2. All preprocessing techniques demonstrated very high validation accuracy, which could mean there is some overfitting, perhaps due to the imbalance of classes.

Table 1: Validation Metrics

| Method | F1 Score | Accuracy | Precision | Recall |
|---|---|---|---|---|
| HOG | 0.991172 | 0.988778 | 0.992995 | 0.989451 |
| CLAHE-HOG | 0.984874 | 0.982147 | 0.986207 | 0.983694 |
| YUV-HOG | 0.994161 | 0.992859 | 0.995757 | 0.992687 |
| HUE-HOG | 0.984368 | 0.978194 | 0.986411 | 0.982435 |
| CLAHE-YUV-HOG | 0.983007 | 0.981127 | 0.983651 | 0.982547 |
| HUE-YUV-HOG | 0.985591 | 0.982020 | 0.987607 | 0.983705 |
| CLAHE-HUE-YUV-HOG | 0.980870 | 0.979214 | 0.981513 | 0.980353 |

Table 2: Test Metrics

| Method | F1 Score | Accuracy | Precision | Recall |
|---|---|---|---|---|
| HOG | 0.879284 | 0.896516 | 0.904510 | 0.864829 |
| CLAHE-HOG | 0.874141 | 0.903325 | 0.885863 | 0.868086 |
| YUV-HOG | **0.890865** | **0.912510** | **0.916218** | **0.876509** |
| HUE-HOG | 0.858395 | 0.880760 | 0.883478 | 0.845495 |
| CLAHE-YUV-HOG | 0.873210 | 0.904909 | 0.887736 | 0.865063 |
| HUE-YUV-HOG | 0.867616 | 0.888519 | 0.893949 | 0.852861 |
| CLAHE-HUE-YUV-HOG | 0.870005 | 0.901267 | 0.884883 | 0.861045 |

Although ultimate performance was not the main goal of this study, but rather compare and identify the most effective preprocessing techniques for enhancing road sign classification accuracy when using a HOG-SVM. Overall, all models achieved relatively high performance around 88% to over 91% accuracy on the test set, and with F1 scores close to 90%. The YUV-HOG model was the only one that managed to beat the baseline HOG in all metrics (91.25% accuracy and 89.09% F1 Score). Adding Gaussian Blur before HOG features generally improved performance in particular the baseline HOG and HUE (about 1-2% overall), except for CLAHE, which showed a slight degradation (about 0.5-1% in test metrics). HUE was nonetheless, the preprocessing step that least performed.

In terms of metrics the F1 score is often preferred in TSR filed because it provides a good balance between precision and recall, particularly in datasets with imbalanced classes, which is the case of our GTSRB dataset. The high accuracy results obtained in this study, particularly with YUV-HOG, demonstrate significant improvement from the baseline HOG-SVM, and not far from the best found in literature. For example, at the time of the benchmark competition a LDA on HOG obtained 95.68%, and was the best HOG model on a board dominated by CNNs such as Arcos-García et al. (2018)'s CNN with 3 Spatial Transformers with 99.71%. Fleyeh and Roch (2013) achieved a classification rate of 99.42% using HOG and Gentle AdaBoost, while Ciresan et al. (2012) achieved 99.46% with deep neural networks on the GTSRB dataset.

In terms of time taken to run the models, initial preprocessing took about 6 to 15 seconds, validation about 1 minute 10 seconds to 1 minute 50 seconds, and testing set slightly less than validation, with YUV-HOG being the fastest overall.

The study was conducted in C++17 using CMAKE in VSCode on a MacBook Pro machine with a 2.3 GHz Dual-Core Intel Core i5 processor and 16GB of RAM.

## 5 Discussion

Combining HOG for feature extraction and SVM for classification proved effective, consistently outputting high validation and test scores for the various preprocessing techniques. Whereas different preprocessing techniques can significantly influence model performance. For instance, YUV-HOG showed considerable improvements in F1 score and accuracy, which confirms its ability to robustly handle diverse light conditions. The inclusion of a Gaussian Blur also led to improved feature extraction, particularly in the HUE-HOG model. Overall all models were able to run and output results in 3 to 5 minutes, and considering the size of the dataset and the machine computing limitations, it is a relatively quick process.

Conversely, a noticeable performance drop from validation to test sets suggests potential overfitting or differences in data distribution and some imbalance in classes leading to difficulties generalising results.

The sequence of preprocessing steps affects results significantly, for example, applying HUE before YUV transformation led to improved performance.

Moreover, combining and stacking multiple preprocessing techniques did not always enhance performance, indicating that more is not necessarily better.

When comparing with other methods such as deep learning CNNs, literature has shown they often outperform SVMs, but the trade off of requiring more computing resources and time. This makes our HOG-SVM preferable for resource and time constraint applications.

There is room for further fine tuning the preprocessing dynamically adjusting to different types of images, as well as a more thorough cross validation, using PCA/LDA and HAAR cascades, or synthetic dataset augmentation of images, such as adding rotated images, balance the image classes, etc. Thus, exploring and optimising the preprocessing pipeline will likely lead to further improvements in the classification, as shown by other researchers.

## 6 Conclusions

This study demonstrates that effective preprocessing techniques can easily and significantly enhance the performance of a HOG-SVM classifier for TSR using the GTSRB dataset. The preprocessing methods like CLAHE and YUV revealed improvements in classification accuracy, YUV increasing the baseline HOG with Gaussian Blur from 89.65% to 91.25%. This improvement confirms the importance of selecting appropriate preprocessing steps to optimise feature extraction in classification problems.

Besides becoming evident that preprocessing is crucial, different techniques showed distinct impacts on model performance. YUV preprocessing, in particular, proved to be highly beneficial, conversely to HUE. Additionally, the sequence in which preprocessing steps are applied can significantly affect the results, as seen with the combination of HUE and YUV transformations. Furthermore, while deep learning models offer high accuracy, traditional methods like HOG-SVM remain preferable for real time and limited resource applications.

Future research should focus on developing dynamic preprocessing methods that adapt based on image characteristics to further enhance classification accuracy. Addressing class imbalance through techniques like synthetic data augmentation could help mitigate overfitting and improve classification. Exploring the integration of advanced techniques such as dimensionality reduction methods (PCA and LDA) and other feature descriptors like HAAR cascades could also lead to improvements. Thorough cross-validation and fine-tuning will be essential to optimize model performance. Finally, ensuring that these improvements are compatible with real time processing constraints is critical for practical ADAS and autonomous vehicle applications.

By addressing these areas, future studies can develop more robust and efficient TSR systems, contributing to safer and more reliable advanced driver assistance systems and autonomous vehicles.